\documentclass{article}
\usepackage{ijcai24}

\usepackage{times}
\usepackage{soul}
\usepackage{url}
\usepackage[hidelinks]{hyperref}
\usepackage[utf8]{inputenc}
\usepackage[small]{caption}
\usepackage{graphicx}
\usepackage{amsmath}
\usepackage{amsthm}
\usepackage{booktabs}
\usepackage{algorithm}
\usepackage{algorithmic}
\usepackage[switch]{lineno}

\title{Self-Knowledge Distillation for Learning Ambiguity}

\author{
Hancheol Park$^1$
\and
Soyeong Jeong$^2$\and
Sukmin Cho$^2$\And
Jong C. Park$^2$\\
\affiliations
$^1$Nota Inc.\\
$^2$School of Computing, KAIST\\
\emails
{hancheol.park}@nota.ai\\
\{starsuzi,nelllpic,jongpark\}@kaist.ac.kr
}
\begin{document}

\maketitle

\begin{abstract}
    Recent language models have shown remarkable performance on natural language understanding (NLU) tasks. However, they are often sub-optimal when faced with ambiguous samples that can be interpreted in multiple ways, over-confidently predicting a single label without consideration for its correctness. To address this issue, we propose a novel self-knowledge distillation method that enables models to learn label distributions more accurately by leveraging knowledge distilled from their lower layers. This approach also includes a learning phase that re-calibrates the unnecessarily strengthened confidence for training samples judged as extremely ambiguous based on the distilled distribution knowledge. We validate our method on diverse NLU benchmark datasets and the experimental results demonstrate its effectiveness in producing better label distributions. Particularly, through the process of re-calibrating the confidence for highly ambiguous samples, the issue of over-confidence when predictions for unseen samples do not match with their ground-truth labels has been significantly alleviated. This has been shown to contribute to generating better distributions than the existing state-of-the-art method. Moreover, our method is more efficient in training the models compared to the existing method, as it does not involve additional training processes to refine label distributions.
\end{abstract}

\section{Introduction}
Natural language understanding (NLU) tasks have been widely employed in various real-life applications. Analyzing sentiments for product reviews is a conventional example of activities aimed at improving the quality of a product ~\cite{toledo-ronen2022}. Recently, it has also become feasible to assess the condition of patients by analyzing the text written by them ~\cite{song2023,wang2023}. However, since there exist a large amount of ambiguous samples, where multiple labels could be correct, in such applications ~\cite{uma2021} (see Table~\ref{tab_1}), models for NLU tasks must carefully make predictions by accurately representing the likelihood of each label. On the contrary, if models predict a single label over-confidently, this could lead to undesirable outcomes. For instance, it may result in a flawed product review analysis and could cause patients to overlook other medical issues, focusing solely on the predicted problems. Unfortunately, it is well-known that state-of-the-art language models for NLU tasks tend to yield over-confident predictions regardless of their incorrectness ~\cite{nie2020,zhang2021,lee2023}.

\begin{table}
\centering
\begin{tabular}{l p{3.8cm}}
\hline
Title & NASA revisiting life on Mars question \\
Ambiguity distribution & Joy: 42\% \newline Surprise: 58\% \newline Others: 0\% \\ 
\hline
Utterance & No. You answer to me whether it is, ``yes'' or ``no.'' Your response? \\
Ambiguity distribution & Attack: 57\% \newline Neutral: 43\% \newline Favor: 0\% \\
\hline
\end{tabular}
\caption{\label{tab_1}
Examples of ambiguous samples from NLU tasks for emotion and question type analysis. ``Ambiguity distribution'' refers to the label distribution obtained from multiple human annotators.}
\end{table}

As an attempt to address this issue, Park and Park ~\shortcite{park2023} have discovered the fact that the degree of relationship between each sample and its labels are more accurately represented in the embedding spaces from lower layers of encoder-based language models, such as BERT ~\cite{devlin2019} and RoBERTa ~\cite{liu2019}. For example, samples annotated only with ``fear'', while expressing both ``fear'' and ``anger'', are positioned very closely to samples annotated with both ``anger'' and ``fear'' in the embedding spaces of lower layers. These samples are distantly located from samples annotated with positive emotions such as ``joy''. However, at the higher layers, such relationships are no longer observed. Based on this observation, they have proposed a method to utilize such relationship information and have demonstrated that output label distributions can be accurately estimated without datasets annotated with empirically-gold ambiguity distributions, which require huge annotation costs. However, this approach involves multiple additional training processes to extract label distributions from lower layers, resulting in high computational complexity for model training. Moreover, this method does not still perfectly outperform the existing methods using datasets annotated with empirically-gold label distributions ~\cite{zhang2018,meissner2021,zhang2021,wang2022}. 

In this work, we propose a novel self-knowledge distillation method to address the issue of training inefficiency and improve the performance of inferring ambiguity distributions. Through the proposed method, models can accurately estimate ambiguity distributions with just a single fine-tuning process. During the fine-tuning process for a target task, the main classifier of the model learns distilled distribution knowledge from a selected lower layer (i.e., a source layer). At the same time, the internal classifier at the source layer learns accurate prediction information from the main classifier to reduce the frequency of conveying incorrect prediction information to the main classifier. Our distillation method also involves the process of re-calibrating that unnecessarily strengthened confidence, which refers to the probability of the predicted label, for samples judged as highly ambiguous based on the distilled distribution information. We expect that this re-calibration process enhances the performance. The self-distillation method that we propose differs from conventional ones that focus on accuracy ~\cite{zhang2019,lee2022,ni2023} in the training processes and the source of knowledge.

Experimental results demonstrate that our method does significantly outperform existing state-of-the-art approaches ~\cite{zhang2021,wang2022,zhou2022,park2023}, in which models are trained using single-labeled datasets. We also observe that the process of re-calibrating the confidence for highly ambiguous training samples effectively addresses the issue of over-confidence when predictions for unseen samples do not match with their ground truth labels. This contributes to reducing the over-confidently predicted probability values for labels other than ground truth labels and increasing the predicted probability values for ground truth labels, thereby improving the label distributions. Furthermore, we demonstrate that, through our proposed self-distillation method, the gradually learned knowledge in the source layer during the fine-tuning process for the target task can be used to accurately learn the ambiguity distributions. This is considerably more efficient compared to the state-of-the-art method ~\cite{park2023} that involves extracting and retraining the distribution knowledge distilled from the source layer after completing fine-tuning for the target task.

\section{Related Work}
Recent language models have shown unprecedented high accuracy on NLU tasks ~\cite{devlin2019,liu2019,lee2023}. Nevertheless, they are poor at estimating label distributions when faced with ambiguous samples ~\cite{pavlick2019,nie2020,lee2023} with the issue of over-confident prediction ~\cite{guo2017}. In this section, we review the existing methods that address this issue on encoder-based pre-trained language models (PLMs), which have been conventionally and popularly used for various NLU tasks.

The most straightforward approach would be to aggregate opinions from multiple annotators for each sample. One can train models with the empirically-gold ambiguity distributions to produce better label distributions for unseen samples ~\cite{zhang2018,meissner2021}. Another approach is using temperature scaling ~\cite{guo2017}, where logits from models fine-tuned for target tasks are re-scaled using a temperature hyperparameter $T$. The $T$ is tuned based on the distribution labels from a validation set ~\cite{zhang2021,wang2022}. Output softmax distributions, in turn, become accordingly closer to gold ambiguity distributions. Although these methods are effective as they directly utilize ground truth distribution information, acquiring accurate distribution labels poses a significant challenge due to the substantial annotation costs involved (e.g., each sample in the ChaosNLI dataset ~\cite{nie2020} is evaluated by 100 annotators). For this reason, most studies have focused on utilizing readily accessible datasets annotated with single labels.

Among the methods over such datasets with single labels, label smoothing ~\cite{muller2019} is known to be easily applicable for learning ambiguity distributions ~\cite{wang2022,zhang2021}. This method softens target one-hot labels by shifting $\alpha$ probability mass from the target labels equally to all the labels. This can naturally help models prevent over-confident predictions. Nevertheless, since this technique does not explicitly consider how to capture the degree of relationship between each sample and its labels, it is difficult to determine that models learn accurate ambiguity distributions through label smoothing. Monte Carlo dropout (MC dropout) ~\cite{gal2016,kendall2017} addresses this by averaging output distributions from $k$ stochastic forward passes, which can be obtained by activating dropout on the inference phase. Since different forward passes may produce possible distributions for ambiguous samples, MC dropout takes into account such relationship information ~\cite{zhou2022}. Nevertheless, this approach has a significant drawback in terms of latency in inference.

Recently, Park and Park ~\shortcite{park2023} address all the issues described above. They extract and use the relationship information between each sample and its labels discovered from lower layers. To identify such relationship information, they use only single-labeled datasets. Their approach outperforms existing methods that use single-labeled training datasets in terms of capturing ambiguity distributions. Nevertheless, there are still several limitations. First, the model is trained multiple times with this approach. More specifically, after fine-tuning the model for the target task, it is necessary to attach internal classifiers to each encoder layer and train them to extract distribution information from lower layers. Then, it is required to retrain the main classifier of the model using distribution information from the source layer. Second, there is still a need for improvement in the performance of inferring ambiguity distributions, because their method does not perfectly outperform methods that use empirically-gold ambiguity distributions. In this work, we propose to resolve such inefficiency problems and further improve performance.

\section{Self-Knowledge Distillation for Learning Ambiguity}
In this section, we describe a novel self-knowledge distillation method and discuss how it can help models to learn ambiguity efficiently and effectively. 
Note that our proposed method is highly efficient, requiring only a single fine-tuning phase for the target task.

We first show that the source layer, which provides accurate distribution knowledge, can be determined at the early stage of fine-tuning (\textsection 3.1). Then, we describe how we mitigate the accuracy degradation issue due to inaccurate prediction information extracted from the source layer during the early stage of training process (\textsection 3.2). Finally, we perform a re-calibration process to reduce the unnecessarily elevated confidence values for training samples with exceptionally high ambiguity, ensuring that highly ambiguous samples do not make over-confident predictions (\textsection 3.3). Such samples can be identified  early in the fined-tuning process (\textsection 3.1).

We design our distillation method based on observations using an emotion analysis dataset ~\cite{mohammad2017}, which is one of the well-kwnown NLU tasks with a significant presence of ambiguous samples. In order to confirm whether our proposed method, designed based on observations from this dataset, can be effectively applied to other datasets, this emotion dataset will not be used in the main experiment.

\subsection{Warm-up Training}
This training phase involves fine-tuning internal classifiers to select the source layer that provides accurate ambiguity distributions to the main classifier of the model. To improve training efficiency, warm-up training is conducted for significantly fewer epochs than those required for entire fine-tuning. Following the previous work ~\cite{park2023}, we assume that the source layer is the layer just before the one in which the most significant drop in the average validation entropy of internal classifiers occurs. The average entropy values are calculated using prediction distributions on a validation set. A significant decrease in entropy indicates a loss of the relationship information between samples and their labels, leading to over-confident predictions.

\begin{figure}
\centering
\includegraphics[draft=false,width=0.8\linewidth]{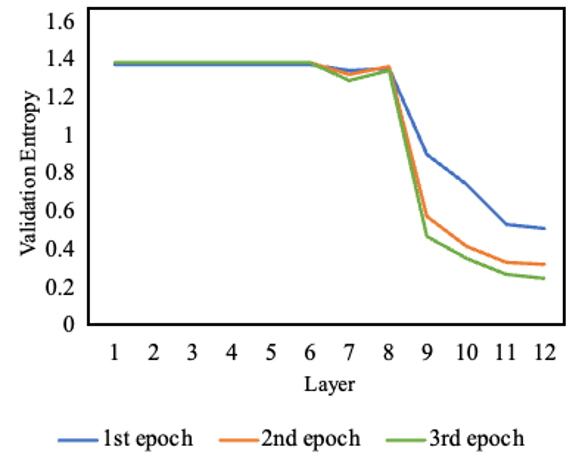}
\caption{Validation entropy in all layers at each epoch}
\label{fig_1}
\end{figure}

More specifically, trainable classifiers are attached to each layer and trained for E epochs with the loss function $\sum_{i=1}^{n} L_i$, where $n$ is the total number of layers and $L_i$ is the cross-entropy loss function with a one-hot label for the $i$-th internal classifier. We define the number of epochs conducted until the selected source layer remains unchanged for at least one epoch as E. For instance, if a layer is identified as the source layer in both the first and second epochs, E is set to 2. The rationale behind this decision is as follows: Through experiments with the emotion analysis dataset, we have observed that the source layer could be identified in the early stage of fine-tuning due to the rapid convergence of PLMs themselves. As illustrated in Figure~\ref{fig_1}, the most significant entropy drop occurs between the 8th and 9th layers after the first epoch, and this pattern persists until the completion of all training (i.e., the 3rd epoch). As we will observe in our subsequent experiments, once the source layer is determined, it remains unchanged (see Discussion section for further details).

In this phase, we also extract m\% of ambiguous samples that are considered to be extremely ambiguous. Such ambiguous samples are determined through the following procedure: After warm-up training is finished, the level of ambiguity (LA) for each sample $x_i$ is computed with the following equation:
\begin{align}
    \resizebox{.91\linewidth}{!}{$
                \displaystyle
    LA(x_i) = \frac{1}{(L - source_{idx} + 1)} *  \sum_{i=source_{idx}}^{L} confidence_{GT}    
    $}
\end{align}%
where $L$ is the total number of layers, $source_{idx}$ is the index of the source layer, and $confidence_{GT}$ is the probability of the ground truth label in the prediction. In the top layer, confidence is high for all samples (due to the issue of over-confidence), but ambiguous samples have lower confidence in the lower layers, resulting in a low $LA$ value. On the other hand, non-ambiguous samples have high confidence values in both higher and lower layers, leading to a high $LA$ value. In this study, We extract samples in ascending order of $LA$ scores, taking the bottom 10\%. This is based on the existing literature, which suggests that most NLU datasets contain at least 10\% of ambiguous samples \cite{uma2021}.

\subsection{Learning Ambiguity Distributions}
After completing the warm-up training, we reset the weights of the model to their initial values. This step is necessary because, 1) during the warm-up phase, the distribution knowledge from the source layer has not yet been transferred to the main classifier (i.e., classifier n), and, 2) depending on the datasets used, there is a possibility that the model could produce over-confident predictions after just a single epoch of training.

\begin{figure}
\centering
\includegraphics[draft=false,width=0.8\linewidth]{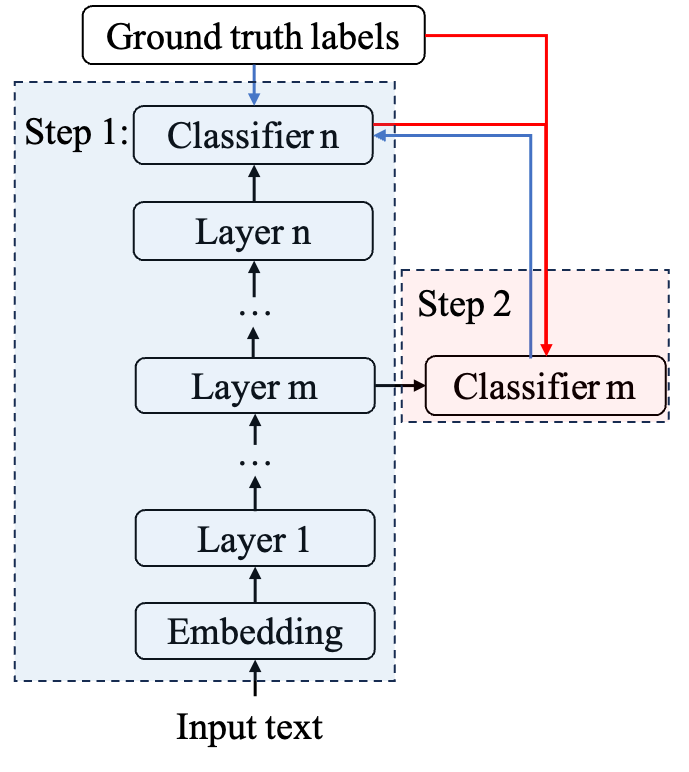}
\caption{The proposed distillation method}
\label{fig_2}
\end{figure}

As illustrated in Figure~\ref{fig_2}, the fine-tuning process for one input batch is conducted in two distinct steps, namely providing accurate ambiguity distribution to the main classifier and accurately predicting information to the source layer. In the first step, the main network (i.e., the network in the box shaded in blue) is trained with the ground truth labels and the distributions distilled from the source layer. The self-distillation loss function for step 1 is defined as follows:
\begin{equation}
    L_{main} = \lambda L_{ce}(\bar{y}_{main}, y) + (1 - \lambda) L_{ce}(\bar{y}_{main}, \bar{y}_{src})
\end{equation}
where $y$ is one-hot label, $\bar{y}_{main}$ is prediction distribution from the main network, $\bar{y}_{src}$ is prediction distribution from the source layer, and $L_{ce}$ is the cross-entropy loss function. We set the value of $\lambda$ to 0.6 to ensure that the main classifier learns information about as many ambiguity distributions as possible from the lower layer, while preventing prediction information that does not match the ground truth labels from overpowering the information about ground truth labels.

\begin{figure}
\centering
\includegraphics[draft=false,width=0.8\linewidth]{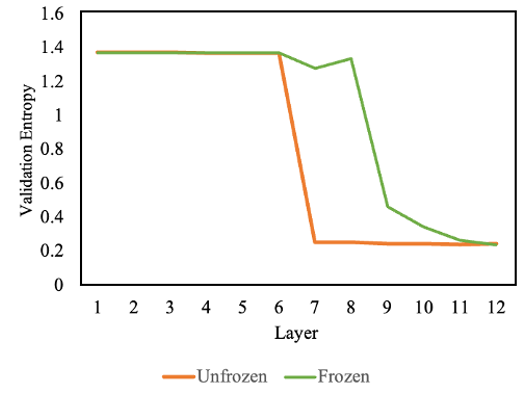}
\caption{The difference of validation entropy between when freezing the weights of the main network and when not freezing them}
\label{fig_3}
\end{figure}

In the second step, the weights of the main network are frozen and only the classifier of the source layer (i.e., classifier m) is fine-tuned with the ground truth labels and the distribution information from the main classifier. This is because we observed that if the main network is not frozen, the relationship information easily disappears in the lower layers, resulting in over-confident predictions in the lower layer (see Figure~\ref{fig_3}). Moreover, in the proposed self-distillation, since the source layer is not sufficiently pre-trained, the main network may receive noise information for a number of training steps, potentially leading to a decrease in accuracy. To address this problem, similar to the traditional self-distillation that aims at improving accuracy, the knowledge from the main network is transferred to the source layer. Accordingly, the self-distillation loss function for step 2 is defined as follows:
\begin{equation}
    L_{src} = \lambda L_{ce}(\bar{y}_{src}, y) + (1 - \lambda) L_{ce}(\bar{y}_{src}, \bar{y}_{main}).
\end{equation}
Here, $\lambda$ is set to 0.6 with the same reasons. 

\subsection{Re-Calibrating Confidences}
After the model learns ambiguity distributions using the distilled knowledge from the source layer, we re-calibrate the unnecessarily strengthened confidences for samples revealed as extremely ambiguous during the warm-up training. Despite the improved accuracy in learning label distributions, we have observed a tendency for the confidences of ambiguous samples to be overestimated. To address this issue, we perform an additional one-epoch training on only the ambiguous sample to adjust their confidence with the following loss function:
\begin{equation}
    \resizebox{.91\linewidth}{!}{$
            \displaystyle
            L_{re-calibration} = 0.5*L_{ce}(\bar{y}_{main}, y) + 0.5*L_{ce}(\bar{y}_{main}, u)
            $}
\end{equation}
where $u$ denotes a uniform distribution. We opt for only one epoch of training because we have observed that as the number of epochs increases, the model tends to predict a uniform distribution for all unseen samples.

\subsection{Comparison}
Note that our proposed self-distillation highly differs from the conventional self-distillation in several aspects. First, our method emphasizes the estimation of ambiguity distributions, rather than solely focusing on enhancing accuracy. Second, we keep the main network frozen during the training of the source layer. Finally, in contrast to traditional approaches that train the entire internal classifiers, our method specifically targets only at the single source layer for training, which makes our method highly efficient.

\section{Experiments}
In this section, we compare our proposed method with existing approaches to validate its effectiveness.

\subsection{Metrics}
We use Jensen-Shannon Distance (JSD) ~\cite{endres2003} and  KL divergence ~\cite{kullback1997,kullback1951} to measure the distance between the prediction distributions of the models and the gold ambiguity distributions. These metrics have been widely used in the previous work ~\cite{nie2020,meissner2021,zhang2021,wang2022,zhou2022}. The KL divergence and JSD are defined by the following equations:

\begin{equation}
    KL({\bf p}||{\bf q}) = \sum_{i \in C} p_i log (\frac{p_i}{q_i})
\end{equation}

\begin{equation}
    JSD({\bf p}||{\bf q}) = \sqrt{\frac{1}{2}(KL({\bf p}||{\bf m}) )+(KL({\bf q}||{\bf m}) )}
\end{equation}

\begin{equation}
    {\bf m} = \frac{1}{2}({\bf p}+{\bf q})
\end{equation}
where ${\bf p}$ is the true ambiguity distribution and ${\bf q}$ is the prediction distribution.

\subsection{Baselines}
As baseline methods, we first compare our method with methods that use the same single gold labels. In this work, we evaluate the performance of ordinary training approach (ORD) (i.e., training using one-hot labels and cross-entropy loss function), MC dropout (MC) ~\cite{zhou2022}, label smoothing (LS) ~\cite{wang2022,zhang2021} and deep model compression (Compression) ~\cite{park2023}. We also employ baselines that utilize additional ambiguity distribution labels such as temperature scaling (TS) ~\cite{wang2022,zhou2022} and label distribution learning (LDL) (i.e., training with ambiguity distributions and cross-entropy loss function) ~\cite{zhang2018,meissner2021}. 

\begin{table*}
\centering
\begin{tabular}{l c c c c c c }
\hline
& \multicolumn{3}{c}{\textbf{ChaosSNLI}} & \multicolumn{3}{c}{\textbf{ChaosMNLI}} \\
& \textbf{JSD$\downarrow$} & \textbf{KL$\downarrow$} & \textbf{Acc.$\uparrow$} 
& \textbf{JSD$\downarrow$} & \textbf{KL$\downarrow$} & \textbf{Acc.$\uparrow$}
\\
\hline
\textbf{ORD}      & 0.3299 & 1.3872 & 0.6935 & 0.4219 & 2.3982 & \textbf{0.5722} \\
\textbf{MC}       & 0.2984 & 0.9287 & 0.6849 & 0.3718 & 1.6320 & 0.5710  \\
\textbf{LS}       & 0.2723 & 0.5724 & 0.7173 & 0.3540 & 0.8574 & 0.5591 \\
\textbf{Compression}     & 0.2635 & 0.3642 & 0.7127 & 0.2799 & 0.4707 & 0.5691 \\
\hline            
\textbf{TS}       & 0.2626 & 0.5099 & 0.6935 & 0.3095 & 0.6491 & \textbf{0.5722} \\
\textbf{LDL}      & \textbf{0.2185} & 0.3811 & 0.7186 & 0.2991 & 0.7032 & 0.5716 \\
\hline
\textbf{LAD}     & 0.2666	& 0.3924 & 0.7008 & 0.2796 & 0.4283 & 0.5660 \\
\textbf{LAD + RC}     & 0.2823 & \textbf{0.3300} & \textbf{0.7272} & \textbf{0.2569} & \textbf{0.3181} & 0.5622 \\
\hline
\end{tabular}
\caption{\label{tab_2}
Performance of existing methods and our proposed method. $\downarrow$ means that a smaller value is better and $\uparrow$ indicates that a bigger value is better. The best values among methods are highlighted. LAD and RC stand for learning ambiguity distributions (Section 3.2) and re-calibration method (Section 3.3), respectively.}
\end{table*}

\begin{table}
\centering
\begin{tabular}{l c c }
\hline
& \textbf{ChaosSNLI} & \textbf{ChaosMNLI} \\
& \textbf{Diff.$\downarrow$} & \textbf{Diff.$\downarrow$} \\
\hline
\textbf{ORD}      & 0.6092 & 0.5749  \\
\textbf{MC}       & 0.5276 & 0.5187  \\
\textbf{LS}       & 0.5753 & 0.5469  \\
\textbf{Compression}     & 0.5265 & 0.4686 \\
\hline            
\textbf{TS}       & 0.5342 & 0.4957  \\
\textbf{LDL}      & 0.4819 & 0.4435  \\
\hline
\textbf{LAD}     & 0.5110 & 0.4479 \\
\textbf{LAD + RC}     & 0.4655 & 0.4122 \\
\hline
\end{tabular}
\caption{\label{tab_3}
The average difference between the ground truth probabilities and predicted probabilities for the ground truth labels when predictions do not match with ground truth labels (Diff.)
}
\end{table}

\subsection{Datasets}
Similar to previous work ~\cite{nie2020,meissner2021,zhang2021,wang2022,zhou2022}, we also use the ChaosNLI dataset ~\cite{nie2020} for evaluating models fine-tuned with NLI tasks. The reason behind this choice is that the ChaosNLI dataset has been considered the most accurate in representing true ground truth ambiguity distributions, as each sample is evaluated by 100 annotators. In the Discussion section, we also validate our method using datasets from other tasks, known for having annotations from many evaluators for each sample, even though they may not be as precise.

Specifically, we used ChaosMNLI (1,599 MNLI-matched development set ~\cite{williams2018}) and ChaosSNLI datasets (1,514 SNLI development set ~\cite{bowman2015}) \cite{nie2020}. For training and validating models, we used AmbiSM datasets \cite{meissner2021}. 
AmbiSM incorporates empirically-gold label distributions gathered through crowd-sourced annotations. It encompasses the SNLI development/test set and the MNLI-matched/mismatched development set, ensuring no sample overlaps with those in ChaosNLI. When assessing models with ChaosMNLI, we designated randomly selected 1,805 MNLI-matched development samples in AmbiSM as the validation set, while the remaining AmbiSM samples (34,395 in total) were used as the training set. For ChaosSNLI, 1,815 SNLI development samples were utilized as the validation set, and the remainder of AmbiSM served as the training set (34,385 samples).

\subsection{Implementation}
All methods in these experiments are implemented based on RoBERTa-base ~\cite{liu2019}. 
All methods utilize identical hyperparameters for training, including a batch size of 32, a learning rate of 5e-5, and a linear decay learning scheduler. The fine-tuning process spans 5 epochs for ChaosSNLI and 7 epochs for ChaosMNLI, as determined by validation accuracy. Keeping the number of training epochs consistent across all methods for each dataset is essential to prevent an increase in overconfident predictions. We employ the AdamW optimizer ~\cite{loshchilov2019}, with a weight decay set to 0.1.

\subsection{Experimental Results}
We report the experimental results that are evaluated on ChaosSNLI and ChaosMNLI datasets in Table \ref{tab_2}. First, we can observe that efficient learning is possible through the knowledge of the source layer, which is progressively trained, without completing fine-tuning for a given task in advance, as in the previous work ~\cite{park2023} . Moreover, using self-distillation together with re-calibrating confidences for ambiguous samples outperforms compression-based method in terms of KL divergence in ChaosSNLI and both of JSD and KL in ChaosMNLI. As shown in Table~\ref{tab_3}, this improvement stems from a significant reduction in the difference between the true probability values of the ground truth labels and the predicted probability values when predictions do not match with ground truth labels. In other words, as the probability values for predicted labels, other than the ground truth label, decrease, there is a natural increase in the prediction probability for the ground truth label. This leads to improvement in the distribution distance between true and prediction distributions. This also indicates that our method mitigates the issue of over-confident prediction when predictions do not match with ground truth labels.

\begin{table*}
\centering
\begin{tabular}{l c c c c c c }
\hline
& \multicolumn{3}{c}{\textbf{ChaosSNLI}} & \multicolumn{3}{c}{\textbf{ChaosMNLI}} \\
& \textbf{JSD$\downarrow$} & \textbf{KL$\downarrow$} & \textbf{Acc.$\uparrow$} 
& \textbf{JSD$\downarrow$} & \textbf{KL$\downarrow$} & \textbf{Acc.$\uparrow$}
\\
\hline
\textbf{Compression}   & 0.2635 & 0.3642 & 0.7127 & 0.2799 & 0.4707 & 0.5691 \\
\textbf{LAD (src $\rightarrow$ main)}   & 0.2597 & 0.3903 & 0.6975 & 0.2815 & 0.4313 & 0.5578 \\
\textbf{+LAD (main $\rightarrow$ src)} & 0.2666 & 0.3924	& 0.7008 & 0.2796 & 0.4283 & 0.5660 \\
\hline
\end{tabular}
\caption{\label{tab_4} Ablation study}
\end{table*}

\section{Discussion}

\begin{figure}
\centering
\includegraphics[draft=false,width=0.8\linewidth]{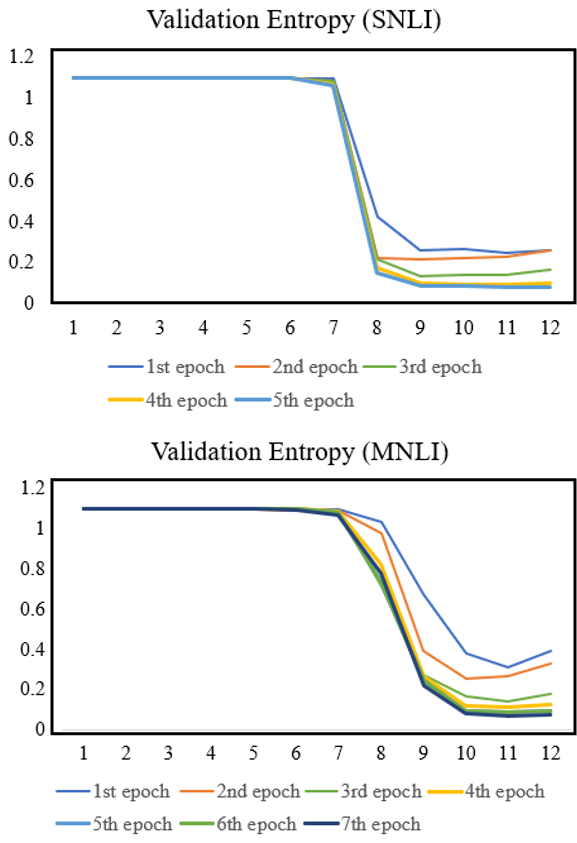}
\caption{Validation entropy in all layers at each epoch}
\label{fig_4}
\end{figure}

\textbf{Do the observations in the warm-up training generally occur?} To answer this question, we investigate the average entropy value for each epoch on the SNLI and MNLI validation sets used in our experiments. As shown in Figure~\ref{fig_4}, a sharp entropy drop occurs in both datasets, and the source layers determined in the first epoch remains consistent until the last training epoch. 

\

\textbf{The effect of self-distillation from the main classifier to the classifier of the source layer} 
As shown in Table~\ref{tab_4}, since the source layer is not trained, the main network receives noise for many training steps, leading to a decrease in accuracy. To address this issue, we improve the accuracy of the source layer by transferring the knowledge from the main network to the source layer to ensure that the improved accuracy of the source layer is reflected properly in the accuracy of the main classifier. As a result, we observed an improvement in accuracy, approaching the performance of the compression-based methods.

\

\begin{figure}
\centering
\includegraphics[draft=false,width=0.7\linewidth]{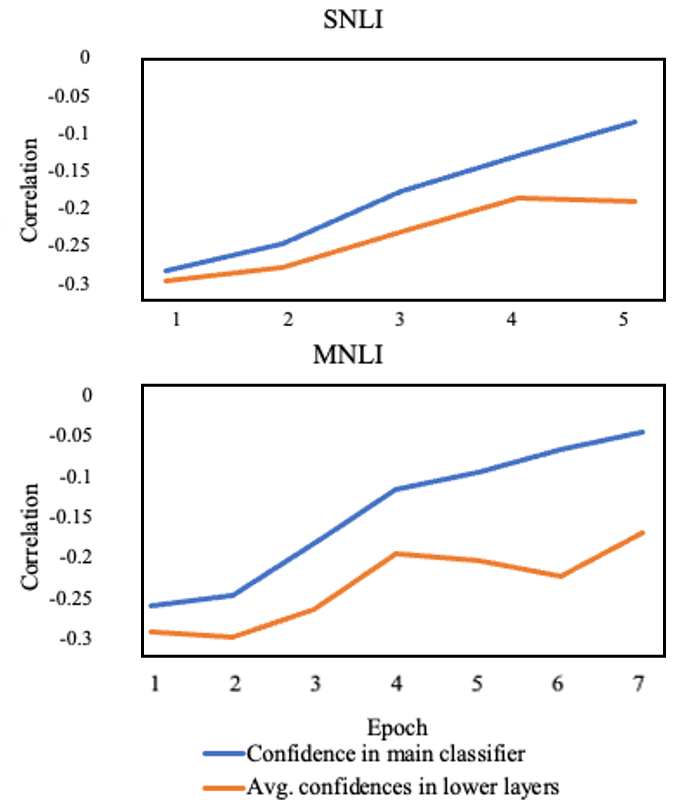}
\caption{Correlation between true ambiguity and predicted ambiguity}
\label{fig_5}
\end{figure}

\textbf{How effective is the level of ambiguity that we propose in selecting ambiguous samples?} It is well-known that the probability of the ground truth label during the fine-tuning process is closely related to the ambiguity of the corresponding sample ~\cite{swayamdipta2020}. To extract ambiguous samples, we calculated the $LA$ by averaging the probabilities of the ground truth label from the source layer to the final layer. To assess the usefulness of our proposed method for selecting ambiguous samples, we computed Pearson correlation coefficient between entropy values of the AmbiNLI training samples and the calculated $LA$ of those samples. As shown in Figure~\ref{fig_5}, our proposed $LA$ exhibits a lower negative correlation (because the high entropy value and the low $LA$ value together indicate that a sample is ambiguous) than the probability of the ground truth label from the main classifier only, especially during the early stages of training. This also suggests the rationality of selecting ambiguous samples during the warm-up training phase.

\

\begin{table*}
\centering
\begin{tabular}{l c c c c c c }
\hline
& \multicolumn{3}{c}{\textbf{ChaosSNLI}} & \multicolumn{3}{c}{\textbf{ChaosMNLI}} \\
& \textbf{JSD$\downarrow$} & \textbf{KL$\downarrow$} & \textbf{Acc.$\uparrow$} 
& \textbf{JSD$\downarrow$} & \textbf{KL$\downarrow$} & \textbf{Acc.$\uparrow$}
\\
\hline
\textbf{LAD + RC (ambiguous samples)}   & 0.2823 & 0.3300 & 0.7272 & 0.2569 & 0.3181 & 0.5622 \\
\textbf{LAD + RC (random samples)} & 0.2902 & 0.3531 & 0.6875 & 0.2611 & 0.3366 & 0.5297 \\
\hline
\end{tabular}
\caption{\label{tab_5} Re-calibrating confidences for ambiguous samples vs. random samples}
\end{table*}

\begin{table}[htb!]
\centering
\begin{tabular}{l c c c c  }
\hline
& \multicolumn{2}{c}{\textbf{Emotion}} & \multicolumn{2}{c}{\textbf{Question Type}} \\
& \textbf{JSD$\downarrow$} & \textbf{KL$\downarrow$} 
& \textbf{JSD$\downarrow$} & \textbf{KL$\downarrow$} 
\\
\hline
\textbf{ORD}      & 0.4203 & 1.2858 & 0.3420 & 1.2248 \\
\textbf{MC}       & 0.4044 & 1.0381 & 0.3140 & 0.9644 \\
\textbf{LS}       & 0.4057 & 0.9825 & 0.3258 & 0.7022 \\
\textbf{Compression} & 0.3935 & 0.8703 & 0.2177 & 0.2898 \\
\hline            
\textbf{TS}       & 0.3859 & 0.7708 & 0.2560 & 0.2769  \\
\textbf{LDL}      & 0.3338 & 0.5198 & 0.1424 & 0.1460  \\
\hline
\textbf{LAD}     & 0.3867 & 0.8602 & 0.1917 & 0.2196  \\
\textbf{LAD + RC}     & 0.3860 & 0.7623 & 0.1941 & 0.1804 \\
\hline
\end{tabular}
\caption{\label{tab_6}
Evaluation results on emotion and question type anaysis tasks}
\end{table}

\textbf{How effective is the use of ambiguous samples during the re-calibration process?} As described in Table~\ref{tab_5}, performing re-calibration for the confidence of random samples results in poorer performance across all performance metrics compared to using ambiguous samples. When using random samples during the re-calibration process, there was a decrease in accuracy. This can be attributed to the fact that using random samples can lead even confident samples to produce ambiguous distributions, causing incorrect predictions due to slight differences in probability values between the correct label and other labels. These results demonstrate the effectiveness of our sampling method.

\

\textbf{Is the proposed method effective on datasets from different tasks?} We also validate the effectiveness of the proposed method on other tasks using emotion ~\cite{strapparava2007} and question type analysis ~\cite{ferracane2021} datasets. For text emotion analysis, we use 800 samples for
training, 200 for validation, and 246 for evaluation from SemEval2007 Task 14 Affective Text dataset ~\cite{strapparava2007}. In this dataset, 6 emotion intensities (i.e., anger, disgust, fear, joy, sadness, and surprise) are annotated by six raters and each intensity value is normalized
to obtain ambiguity distributions using the same procedure as in the previous work ~\cite{zhang2018}.
For question type analysis, we use a discourse dataset ~\cite{ferracane2021}. We use 720 samples for training, 80 for validation, and 200 for evaluation. This dataset contains three labels for question types such as ``attack'', ``favor'', and ``neutral''. We obtain ambiguity distributions by normalizing the results from seven annotations. As shown in Table~\ref{tab_6}, we obtained similar results in our experiments using the ChaosNLI dataset, even though these datasets are less precise in representing distribution information compared to the ChaosNLI dataset~\footnote{The number of test samples is not large, and there is almost no difference in the count of mis-predicted samples among methods. Therefore, accuracy was not separately reported.}.

\section{Conclusion}
In this work, we have proposed a novel self-knowledge distillation method that helps models learn ambiguity more efficiently and effectively. We also demonstrate the effectiveness of our method on various NLU benchmarks. Unlike exiting studies that have been only evaluated on NLI tasks, we have validated our method across a broader range of domains. 

In this work, we focused on the encoder-based language model, which is widely used in many classification tasks. Recently, however, decoder-based large language models (LLMs) with billions number of parameters have been applied to several NLU tasks. Despite their growing usage, the ability of these decoder-based LLMs to capture ambiguity remains relatively understudied. Although it is beyond the scope of our current work, we believe that it would be a meaningful future research direction to investigate how decoder-based LLMs handle ambiguity and further explore whether the knowledge from the lower layer can also be utilized, especially given their vast parameter space and sophisticated architecture.

\bibliographystyle{named}
\bibliography{ijcai24}

\begin{thebibliography}{}

\bibitem[\protect\citeauthoryear{Bowman \bgroup \em et al.\egroup }{2015}]{bowman2015}
Samuel~R. Bowman, Gabor Angeli, Christopher Potts, and Christopher~D. Manning.
\newblock A large annotated corpus for learning natural language inference.
\newblock In {\em Proceedings of the 2015 Conference on Empirical Methods in Natural Language Processing}, pages 632--642, Lisbon, Portugal, September 2015. Association for Computational Linguistics.

\bibitem[\protect\citeauthoryear{Devlin \bgroup \em et al.\egroup }{2019}]{devlin2019}
Jacob Devlin, Ming-Wei Chang, Kenton Lee, and Kristina Toutanova.
\newblock Bert: Pre-training of deep bidirectional transformers for language understanding.
\newblock In {\em Proceedings of the 2019 Conference of the North American Chapter of the Association for Computational Linguistics: Human Language Technologies, Volume 1 (Long and Short Papers)}, pages 4171--4186, Minneapolis, Minnesota, June 2019. Association for Computational Linguistics.

\bibitem[\protect\citeauthoryear{Endres and Schindelin}{2003}]{endres2003}
Dominik~Maria Endres and Johannes~E. Schindelin.
\newblock A new metric for probability distributions.
\newblock {\em IEEE Transactions on Information Theory}, 49(7):1858--1860, June 2003.

\bibitem[\protect\citeauthoryear{Ferracane \bgroup \em et al.\egroup }{2021}]{ferracane2021}
Elisa Ferracane, Greg Durrett, Junyi~Jessy Li, and Katrin Erk.
\newblock Did they answer? subjective acts and intents in conversational discourse.
\newblock In {\em Proceedings of the 2021 Conference of the North American Chapter of the Association for Computational Linguistics: Human Language Technologies}, pages 1626--1644, Online, June 2021. Association for Computational Linguistics.

\bibitem[\protect\citeauthoryear{Gal and Ghahramani}{2016}]{gal2016}
Yarin Gal and Zoubin Ghahramani.
\newblock Dropout as a bayesian approximation: Representing model uncertainty in deep learning.
\newblock In {\em Proceedings of The 33rd International Conference on Machine Learning}, pages 1050--1059, 2016.

\bibitem[\protect\citeauthoryear{Guo \bgroup \em et al.\egroup }{2017}]{guo2017}
Chuan Guo, Geoff Pleiss, Yu~Sun, and Kilian~Q. Weinberger.
\newblock On calibration of modern neural networks.
\newblock In {\em Proceedings of The 34th International Conference on Machine Learning}, pages 1321--1330, 2017.

\bibitem[\protect\citeauthoryear{Kendall and Gal}{2017}]{kendall2017}
Alex Kendall and Yarin Gal.
\newblock What uncertainties do we need in bayesian deep learning for computer vision?
\newblock In {\em Proceedings of Advances in Neural Information Processing Systems 31 (NeurIPS 2017)}, 2017.

\bibitem[\protect\citeauthoryear{Kullback and Leibler}{1951}]{kullback1951}
S.~Kullback and R.~A. Leibler.
\newblock On information and sufficiency.
\newblock {\em Transactions of the Association for Computational Linguistics}, 22(1):79--86, 1951.

\bibitem[\protect\citeauthoryear{Kullback}{1997}]{kullback1997}
S.~Kullback.
\newblock {\em Information theory and statistics}.
\newblock Courier Corporation, 1997.

\bibitem[\protect\citeauthoryear{Lee and Lee}{2022}]{lee2022}
Hojung Lee and Jong-Seok Lee.
\newblock Rethinking online knowledge distillation with multi-exits.
\newblock In {\em Proceedings of the Asian Conference on Computer Vision}, 2022.

\bibitem[\protect\citeauthoryear{Lee \bgroup \em et al.\egroup }{2023}]{lee2023}
Noah Lee, Na~Min An, and James Thorne.
\newblock Can large language models capture dissenting human voices?
\newblock In {\em Proceedings of the 2023 Conference on Empirical Methods in Natural Language Processing}, pages 4569--4585, Singapore, December 2023. Association for Computational Linguistics.

\bibitem[\protect\citeauthoryear{Liu \bgroup \em et al.\egroup }{2019}]{liu2019}
Yinhan Liu, Myle Ott, Naman Goyal, Jingfei Du, Mandar Joshi, Danqi Chen, Omer Levy, Mike Lewis, Luke Zettlemoyer, and Veselin Stoyanov.
\newblock Roberta: A robustly optimized bert pretraining approach.
\newblock {\em arXiv preprint arXiv:1907.11692}, 2019.

\bibitem[\protect\citeauthoryear{Loshchilov and Hutter}{2019}]{loshchilov2019}
Ilya Loshchilov and Frank Hutter.
\newblock Decoupled weight decay regularization.
\newblock In {\em Proceedings of International Conference on Learning Representations}, 2019.

\bibitem[\protect\citeauthoryear{Meissner \bgroup \em et al.\egroup }{2021}]{meissner2021}
Johannes~Mario Meissner, Napat Thumwanit, Saku Sugawara, and Akiko Aizawa.
\newblock Embracing ambiguity: Shifting the training target of nli models.
\newblock In {\em Proceedings of the 59th Annual Meeting of the Association for Computational Linguistics and the 11th International Joint Conference on Natural Language Processing (Volume 2: Short Papers)}, pages 862--869, Online, August 2021. Association for Computational Linguistics.

\bibitem[\protect\citeauthoryear{Mohammad and Bravo-Marquez}{2017}]{mohammad2017}
Saif Mohammad and Felipe Bravo-Marquez.
\newblock Emotion intensities in tweets.
\newblock In {\em Proceedings of the 6th Joint Conference on Lexical and Computational Semantics (*SEM 2017)}, pages 65--77, Vancouver, Canada, August 2017. Association for Computational Linguistics.

\bibitem[\protect\citeauthoryear{Müller \bgroup \em et al.\egroup }{2019}]{muller2019}
Rafael Müller, Simon Kornblith, and Geoffrey Hinton.
\newblock When does label smoothing help?
\newblock In {\em Proceedings of the 33rd International Conference on Neural Information Processing Systems}, pages 4694--4703, 2019.

\bibitem[\protect\citeauthoryear{Ni \bgroup \em et al.\egroup }{2023}]{ni2023}
Shuiping Ni, Xinliang Ma, Mingfu Zhu, Xingwang Li, and Yu-Dong Zhang.
\newblock Reverse self-distillation overcoming the self-distillation barrier.
\newblock {\em IEEE Open Journal of the Computer Society}, 2023.

\bibitem[\protect\citeauthoryear{Nie \bgroup \em et al.\egroup }{2020}]{nie2020}
Yixin Nie, Xiang Zhou, and Mohit Bansal.
\newblock What can we learn from collective human opinions on natural language inference data?
\newblock In {\em Proceedings of the 2020 Conference on Empirical Methods in Natural Language Processing (EMNLP)}, pages 9131--9143, Online, November 2020. Association for Computational Linguistics.

\bibitem[\protect\citeauthoryear{Park and Park}{2023}]{park2023}
Hancheol Park and Jong Park.
\newblock Deep model compression also helps models capture ambiguity.
\newblock In {\em Proceedings of the 61st Annual Meeting of the Association for Computational Linguistics (Volume 1: Long Papers)}, pages 6893--6905, Toronto, Canada, July 2023. Association for Computational Linguistics.

\bibitem[\protect\citeauthoryear{Pavlick and Kwiatkowski}{2019}]{pavlick2019}
Ellie Pavlick and Tom Kwiatkowski.
\newblock Inherent disagreements in human textual inferences.
\newblock {\em Transactions of the Association for Computational Linguistics}, 7:677--694, 2019.

\bibitem[\protect\citeauthoryear{Song \bgroup \em et al.\egroup }{2023}]{song2023}
Hoyun Song, Jisu Shin, Huije Lee, and Jong Park.
\newblock A simple and flexible modeling for mental disorder detection by learning from clinical questionnaires.
\newblock In {\em Proceedings of the 61st Annual Meeting of the Association for Computational Linguistics (Volume 1: Long Papers)}, pages 12190--12206, Toronto, Canada, July 2023. Association for Computational Linguistics.

\bibitem[\protect\citeauthoryear{Strapparava and Mihalcea}{2007}]{strapparava2007}
Carlo Strapparava and Rada Mihalcea.
\newblock Semeval-2007 task 14: Affective text.
\newblock In {\em Proceedings of the Fourth International Workshop on Semantic Evaluations (SemEval-2007)}, pages 70--74, Prague, Czech Republic, June 2007. Association for Computational Linguistics.

\bibitem[\protect\citeauthoryear{Swayamdipta \bgroup \em et al.\egroup }{2020}]{swayamdipta2020}
Swabha Swayamdipta, Roy Schwartz, Nicholas Lourie, Yizhong Wang, Hannaneh Hajishirzi, Noah~A. Smith, and Yejin Choi.
\newblock Dataset cartography: Mapping and diagnosing datasets with training dynamics.
\newblock In {\em Proceedings of the 2020 Conference on Empirical Methods in Natural Language Processing (EMNLP)}, pages 9275--9293, Online, November 2020. Association for Computational Linguistics.

\bibitem[\protect\citeauthoryear{Toledo-Ronen \bgroup \em et al.\egroup }{2022}]{toledo-ronen2022}
Orith Toledo-Ronen, Matan Orbach, Yoav Katz, and Noam Slonim.
\newblock Multi-domain targeted sentiment analysis.
\newblock In {\em Proceedings of the 2022 Conference of the North American Chapter of the Association for Computational Linguistics: Human Language Technologies}, pages 2751--2762, Seattle, United States, July 2022. Association for Computational Linguistics.

\bibitem[\protect\citeauthoryear{Uma \bgroup \em et al.\egroup }{2021}]{uma2021}
Alexandra~N. Uma, Tommaso Fornaciari, Dirk Hovy, Silviu Paun, Barbara Plank, and Massimo Poesio.
\newblock Learning from disagreement: A survey.
\newblock {\em Journal of Artificial Intelligence Research}, 72:1385--1470, 2021.

\bibitem[\protect\citeauthoryear{Wang \bgroup \em et al.\egroup }{2022}]{wang2022}
Yuxia Wang, Minghan Wang, Yimeng Chen, Shimin Tao, Jiaxin Guo, Chang Su, Min Zhang, and Hao Yang.
\newblock Capture human disagreement distributions by calibrated networks for natural language inference.
\newblock In {\em Findings of the Association for Computational Linguistics: ACL 2022}, pages 1524--1535, Dublin, Ireland, May 2022. Association for Computational Linguistics.

\bibitem[\protect\citeauthoryear{Wang \bgroup \em et al.\egroup }{2023}]{wang2023}
Benyou Wang, Qianqian Xie, Jiahuan Pei, Zhihong Chen, Prayag Tiwari, Zhao Li, and Jie fu.
\newblock Pre-trained language models in biomedical domain.
\newblock {\em ACM Computing Surveys}, 56, 2023.

\bibitem[\protect\citeauthoryear{Williams \bgroup \em et al.\egroup }{2018}]{williams2018}
Adina Williams, Nikita Nangia, and Samuel Bowman.
\newblock A broad-coverage challenge corpus for sentence understanding through inference.
\newblock In {\em Proceedings of the 2018 Conference of the North American Chapter of the Association for Computational Linguistics: Human Language Technologies, Volume 1 (Long Papers)}, pages 1112--1122, New Orleans, Louisiana, June 2018. Association for Computational Linguistics.

\bibitem[\protect\citeauthoryear{Zhang \bgroup \em et al.\egroup }{2018}]{zhang2018}
Yuxiang Zhang, Jiamei Fu, Dongyu She, Ying Zhang, Senzhang Wang, and Jufeng Yang.
\newblock Text emotion distribution learning via multi-task convolutional neural network.
\newblock In {\em Proceedings of the Twenty-Seventh International Joint Conference on Artificial Intelligence (IJCAI)}, pages 4595--4601, 2018.

\bibitem[\protect\citeauthoryear{Zhang \bgroup \em et al.\egroup }{2019}]{zhang2019}
Linfeng Zhang, Jiebo Song, Anni Gao, Jingwei Chen, Chenglong Bao, and Kaisheng Ma.
\newblock Be your own teacher: Improve the performance of convolutional neural networks via self distillation.
\newblock In {\em Proceedings of the IEEE/CVF International Conference on Computer Vision}, pages 3713--3722, 2019.

\bibitem[\protect\citeauthoryear{Zhang \bgroup \em et al.\egroup }{2021}]{zhang2021}
Shujian Zhang, Chengyue Gong, and Eunsol Choi.
\newblock Learning with different amounts of annotation: From zero to many labels.
\newblock In {\em Proceedings of the 2021 Conference on Empirical Methods in Natural Language Processing}, pages 7620--7632, Online, November 2021. Association for Computational Linguistics.

\bibitem[\protect\citeauthoryear{Zhou \bgroup \em et al.\egroup }{2022}]{zhou2022}
Xiang Zhou, Yixin Nie, and Mohit Bansal.
\newblock Distributed nli: Learning to predict human opinion distributions for language reasoning.
\newblock In {\em Findings of the Association for Computational Linguistics: ACL 2022}, pages 972--987, Dublin, Ireland, May 2022. Association for Computational Linguistics.

\end{thebibliography}

\end{document}